\documentclass[10pt,twocolumn,letterpaper]{article}

\usepackage{cvpr}
\usepackage{times}
\usepackage{epsfig}
\usepackage{graphicx}
\usepackage{amsmath}
\usepackage{amssymb}
\usepackage{multirow}
\usepackage{array}
\usepackage{lscape}
\usepackage{booktabs}
\usepackage{threeparttable}

\usepackage[breaklinks=true,bookmarks=false]{hyperref}

\cvprfinalcopy 

\def\cvprPaperID{****} 

\begin{document}

%

	\title{Investigations on the inference optimization techniques and their impact on multiple hardware platforms for Semantic Segmentation}
	\author{Sethu Hareesh Kolluru\\
    {\tt\small hareesh@stanford.edu}\\
    }
	\maketitle


\begin{abstract}
In this work, the task of pixel-wise semantic segmentation in the context of self-driving with a goal to reduce the inference time is explored. Fully Convolutional Network (FCN-8s, FCN-16s and FCN-32s) with a VGG16 encoder architecture and skip connections is trained and validated on Cityscapes dataset. Numerical investigations are carried out for several inference optimization techniques built into TensorFlow and TensorRT to quantify their impact on the inference time and network size. Finally, the trained network is ported on to an embedded platform (Nvidia Jetson TX1) and inference time as well as total energy consumed for inference across hardware platforms are compared.
\end{abstract}



\section{Background and Motivation}
Semantic segmentation is the ability to understand an image at the pixel level and assigning a label from a group of classes to every pixel. Semantic segmentation comes in two flavors, one that does not differentiate between object instances of the same class, referred to as pixel-level semantic segmentation and one that does, instance-level semantic segmentation. An example of an image that has been semantically labeled with objects of different classes such as roads, people, trees, etc shown in different colors is shown below.
\begin{figure}[!hbt]

		\begin{center}
		\includegraphics[width=\linewidth]{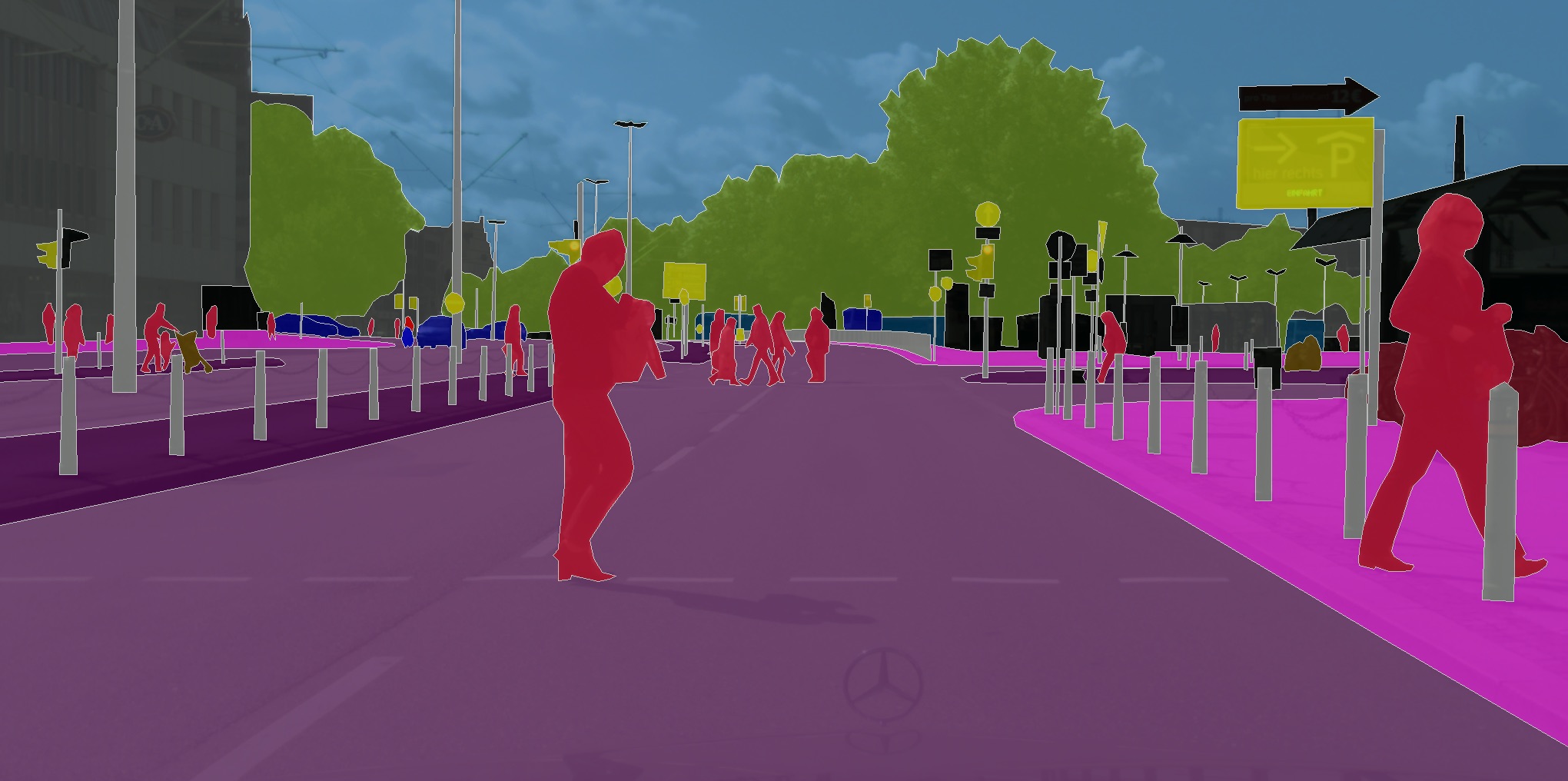}
		\caption{An annotated example image from Cityscapes dataset}
		\label{fig:tf_plot}
		\end{center}
	\end{figure}
     
A popular application of semantic segmentation is in autonomous driving systems, 
where reliable and accurate scene understanding is a critical component. In addition, there is also a strong requirement to segment the image in real-time as the self-driving car needs to react instantly to new events to guarantee the safety of the personnel involved \cite{speeding}.

The advent of deep learning has made great strides towards better visual understanding \cite{nature} and in particular, semantic segmentation; However, this performance was accomplished by increasing the depth of network as well as the computational infrastructure required. This poses even a greater challenge in the context of self-driving, as deploying such deep networks that work as inference engines is not feasible or at least difficult in an embedded device with a limited compute capability in a self-driving car.

Hence, there is a need to quantify and understand the network's end-to-end response time i.e inference time, the bottlenecks that dictate it as well as methods or techniques that can be employed to improve it. 

In this work, a Fully Convolutional Network architecture for the task of pixel-wise semantic segmentation on Cityscapes dataset is implemented and performance metrics are obtained. Numerical investigations are carried out for several inference optimization techniques such as weight quantization with a goal towards improving inference time. Finally, the trained model is then ported to an embedded platform  (Nvidia Jetson TX1) and inference times are quantified when built-in optimizations in Nvidia's TensorRT inference engine are enabled. 






\section{Related Work}
In this section, literature work related to semantic segmentation, its application in the field of self-driving and the enablers (datasets) and the corresponding challenges in this context are outlined into three subcategories.

\subsection{Deep semantic segmentation}
Semantic segmentation, which was viewed as a challenging problem in computer vision until a few years ago, has witnessed rapid progress recently with deep learning\cite{deepseg}. One of the seminal works in this area that brought focus on the end-to-end learning of pixel-wise classification is the Fully Convolutional Network (FCN) architecture, which does not have any fully-connected layers at the end, that is typically used for classification but instead employs convolutional layers to classify each pixel in the image \cite{long}. The key insight in this work is that, the network first learns feature maps, whose height and width dimensions, are reduced by striding and pooling operations; which are then upsampled within the network using transpose convolution (or deconvolution), so that dimensions of output match that of the original input image, to get dense predictions. 

One of the principal limitations of this approach, however, is the impact of the loss of resolution on the final prediction as the architecture relies on first downsampling the image into feature maps. This is addressed in \cite{noh}, where a deeper transpose convolution network, with stacked deconvolution layers and unpooling layers, was employed to achieve performance gain as the deconvolution network is overly simple and the input to it is too coarse in \cite{long}. In SegNet\cite{segnet}, a similar approach with an encoder-decoder architecture is used to address the loss of detailed structures of an object due to a coarse feature map; The decoder network, however, uses the maxpooling indices from the corresponding encode layer to perform upsampling. 

The issue of multi-scale semantics is the focus in \cite{unet}, \cite{yu}. Networks that work with a fixed size receptive field, can only handle single scale semantics. i.e if the object is substantially larger or smaller than the receptive field, then it is either fragmented or mislabeled. Building upon the idea of skip-architecture as proposed in \cite{long} to merge feature maps from different resolutions, U-Net\cite{unet}, a U-shaped encoder-decoder architecture network is developed, where feature maps from different initial layers are upsampled and added for the next layers is developed. Another work by \cite{yu} introduced dilated convolutions to aggressively increase the receptive field of the kernel without introducing parameters or subsampling, which provided a better solution for handling multiple scales. 

\subsection{Semantic segmentation in self-driving}
Shifting gears into the application of semantic segmentation in the context of scene understanding in self-driving systems that puts forth the need to reduce the inference latency and hence the computation required. One approach is to come up with computationally efficient architectures such as Squeezenet\cite{squeezenet}, which demonstrated that it is possible to reproduce the image classification accuracy of Alexnet\cite{alexnet} using 50x fewer parameters by using a more efficient architecture. ENet\cite{enet} also presented a more efficient architecture with convolutional layer factorization. This is achieved by decomposing each $n$ x $n$ convolution into two smaller ones following each other: one with a $n$ x $1$ filter and the other with a $1$ x $n$ filter, which allows for large speedups, and greatly reduces the number of parameters, thus, making them less redundant. 

Another line of research focuses on increasing the efficiency of existing networks by deriving smaller networks from larger counterparts \cite{hinton}, or by pruning or quantizing weights \cite{han}. Another trend in the industry is to tweak the network for execution on specific hardware design or implement them using platform specific libraries such as TensorRT that optimizes deep learning models for inference and creates a runtime for deployment on specific hardware platforms. 

\subsection{Datasets for street scene understanding}
A major contributing factor to the progress of deep learning, especially to the problem of image classification is the availability of large-scale, publicly-available datasets such as ImageNet\cite{imagenet}. 
Similarly, research progress in the application of semantic segmentation in self-driving for street scene understanding can be related to the existence of datasets such as KITTI Vision benchmark suite \cite{kitti} and Camdvid \cite{camvid}. However, these datasets are relatively smaller and do not fully capture the variability and complexity of real world scenarios. Cityscapes is a high quality dataset for semantic street scene understanding with labeled examples of actual road scene images from 50 German cities collected in different weather conditions and, therefore, is tailored for autonomous driving in an urban environment \cite{cordts}. A more recent effort to build a much larger dataset resulted in the Mapillary Vistas dataset \cite{mapillary}, with $25,000$ images with $100$ classes. 

Despite the significant progress, the best inference times when it comes to semantic segmentation task in the embedded system are still less than $5$ frames per second \cite{enet}, \cite{speeding}, which is clearly not acceptable as a viable commercial solution. Hence, there is a need to build upon these ideas and explore methods to reduce interference time.

\section{Architecture}
FCN architecture, which still serves as a blueprint for most segmentation architectures, is employed in this study\cite{long}. The network is composed of two parts - encoder and decoder. The encoder network corresponds to the feature extractor that transforms the input image to a multidimensional feature representation, whereas the decoder derives the semantic segmentation map from the features extracted from the encoder network. 

\subsection{Encoder}

\begin{figure}[!hbt]

		\begin{center}
		\includegraphics[width=\linewidth]{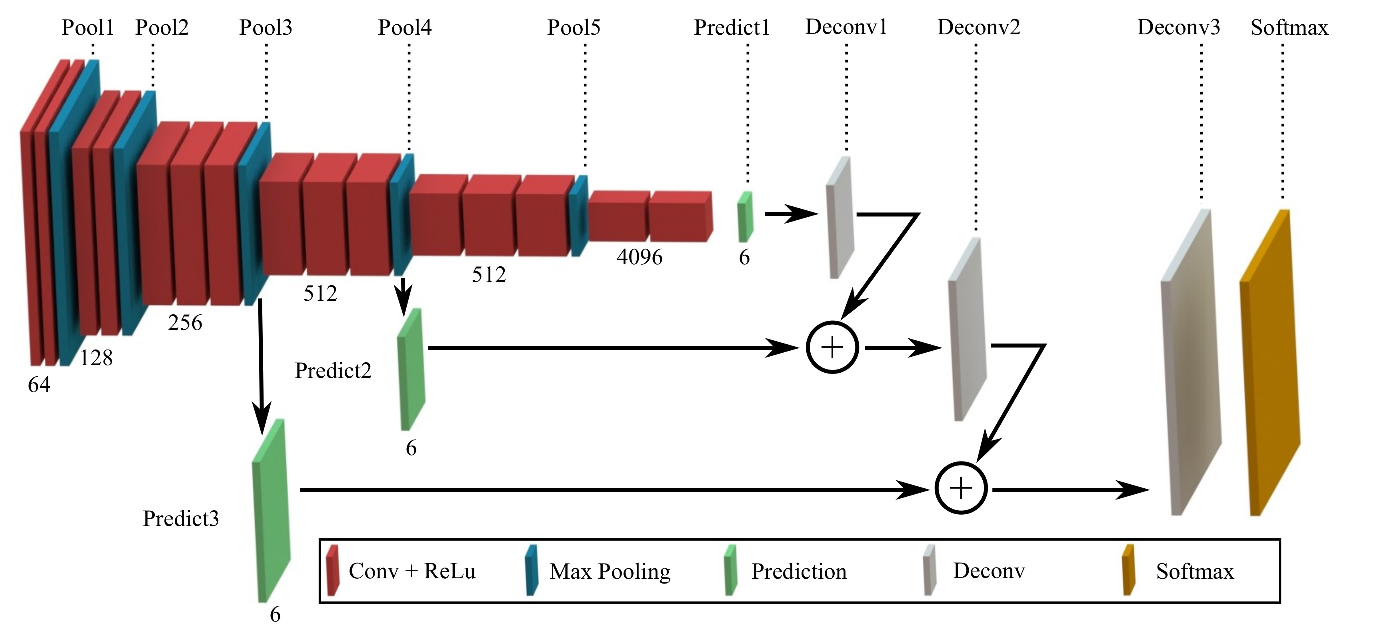}
		\caption{Network Architecture: FCN-32, FCN-16, FCN-8}
		\label{fig:Encoder }
		\end{center}
	\end{figure}

The encoder is a modified VGG16 architecture \cite{vgg}, which was initially designed for the task of image classification and has been shown to generalize well to other datasets and a popular encoder choice for segmentation task as well. The main contribution of this architecture is its use of very small (3x3) convolution filters. It has demonstrated that replacing large kernel-sized filters with multiple 3x3 kernel-sized filters one after another for a given receptive field (the effective area size of input image on which output depends), 
enables it to learn more complex features, and that too at a lower cost. 

This network is modified by replacing the three fully connected layers at the end with three 1x1 convolution layers with $4096$, $4096$ and $35$ (the number of classes in the dataset) number of filters respectively.

\begin{table}[!hbt]
\small
\begin{tabular}{|@{}c@{}|@{}c@{}|@{}c@{}|}
\hline
\begin{tabular}{@{}c@{}}Input\\ $(256\times512\times3)$\end{tabular} &  Activations &  Parameters \\
\hline
\hline
Conv$1_1$ : 3x3-64	& $256\times512\times64$ &	$(3\times3\times3 + 1)\times64$\\
\hline 
Conv$1_2$ : 3x3-64	& $256\times512\times64$ &	$(3\times3\times3 + 1)\times64$\\
\hline 
Pool$1$	& $128\times256\times64$ &	0\\
\hline 
Conv$2_1$ : 3x3-128	& $128\times256\times128$ &	$(3\times3\times64 + 1)\times128$\\
\hline 
Conv$2_2$ : 3x3-128	& $128\times256\times128$ &	$(3\times3\times64 + 1)\times128$\\
\hline 
Pool$2$	& $64\times128\times128$ &	0\\
\hline 
Conv$3_1$ : 3x3-256	& $64\times128\times256$ &	$(3\times3\times128+ 1)\times256$\\
\hline 
Conv$3_2$ : 3x3-256	& $64\times128\times256$ &	$(3\times3\times128+ 1)\times256$\\
\hline
Conv$3_3$ : 3x3-256	& $64\times128\times256 $&	$(3\times3\times128+ 1)\times256$\\
\hline 
Pool$4$	& $32\times64\times256$ &	0\\
\hline 
Conv$4_1$ : 3x3-512	& $32\times64\times512$ &	$(3\times3\times256+ 1)\times512$\\
\hline 
Conv$4_2$ : 3x3-512	& $32\times64\times512$ &	$(3\times3\times256+ 1)\times512$\\
\hline
Conv$4_3$ : 3x3-512	& $32\times64\times512$ &	$(3\times3\times256+ 1)\times512$\\
\hline 
Pool$4$	& $16\times32\times512$ &	0\\
\hline 
Conv$5_1$ : 3x3-512	& $16\times32\times512$ &	$(3\times3\times512+ 1)\times512$\\
\hline 
Conv$5_2$ : 3x3-512	& $16\times32\times512$ &	$(3\times3\times512+ 1)\times512$\\
\hline
Conv$5_3$ : 3x3-512	& $16\times32\times512$ &	$(3\times3\times512+ 1)\times512$\\
\hline 
Pool$5$	& $8\times16\times512$ &	0\\
\hline 
Conv$6$ : 1x1-4096	& $8\times16\times4096$ &	$(1\times1\times512 + 1)\times4096$\\
\hline 
Conv$7$ : 1x1-4096	& $8\times16\times4096$ &	$(1\times1\times4096 + 1)\times4096$\\
\hline
Conv$1by1$ : 1x1-35	& $8\times16\times35$ &	$(1\times1\times4096 + 1)\times35$\\
\hline
\hline
Total Memory & ~154 MB & ~128 MB \\
\hline
\end{tabular}	
\caption{Encoder Memory Estimates: Activations and Parameters}
\label{tab:encoder_memory}
\end{table}

\begin{table}[!hbt]
\small
\begin{center}
\begin{tabular}{|@{}c@{}|@{}c@{}|@{}c@{}|}
\hline
\begin{tabular}{@{}c@{}}Input from Encoder \\ (8 x 16 x 35)\end{tabular} &  Activations  &  Parameters \\
\hline
\hline
up4: conv-transpose	& 16 x 32 x 512 &	(4x4x35 + 1) x 512\\
\hline 
skip4: add	& (16 x 32 x 512) x 2 &	0\\
\hline 
up3: conv-transpose	& 32 x 64 x 256 &	(4 x 4 x 512 + 1) x 256\\
\hline 
skip3 : add	& (32 x 64 x 256) x 2 &	0\\
\hline 
output:conv-transpose	& 256 x 512 x 35 &	(16 x 16 x 256 + 1) x 35\\
\hline 
\hline
Total Memory & ~17.5 MB & ~8.75 MB \\
\hline
\end{tabular}
\caption{Decoder(FCN-8s) Memory Estimates : Activations and Parameters}
\label{tab:decoder_memory}
\end{center}
\end{table}

\subsection{Decoder}
\begin{figure}[!hbt]

		\begin{center}
		\includegraphics[width=\linewidth]{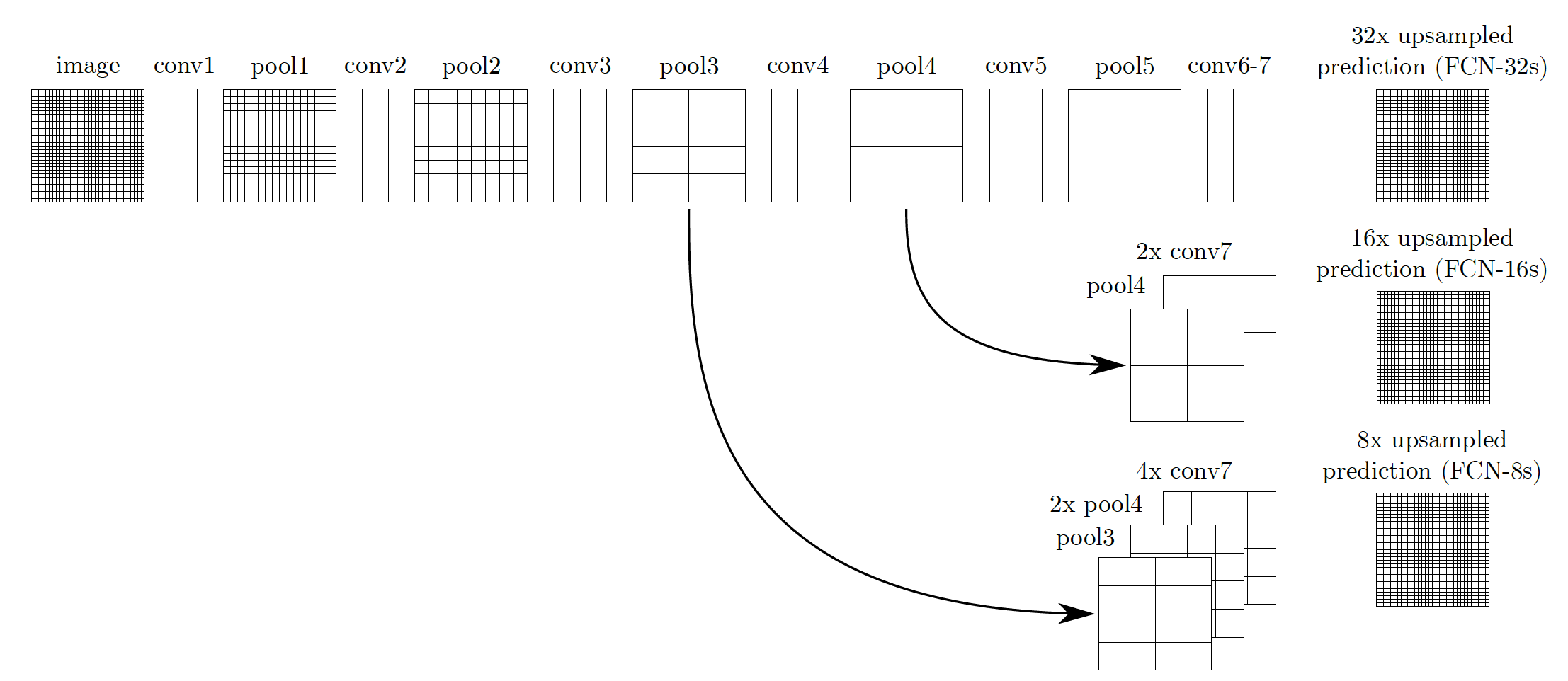}
		\caption{Decoder Architecture: FCN-32s, FCN-16s, FCN-8s}
		\label{fig:Decoder }
		\end{center}
	\end{figure}

The task of semantic segmentation can be interpreted as understanding ``what" class a pixel belongs to as well as ``where" the pixel is in the original image\cite{long}. The challenge, however, is that the semantic information resides in the deeper layers, which are coarser in resolution, while the location information resides in the shallower layers, which are finer in resolution. Therefore, to improve dense prediction, coarse, semantic information from deeper layers is combined with finer, appearance information and the way in which they are fused together results in different decoder architectures. In particular, three different networks are used in this study - FCN-32s, FCN-16s, and FCN-8s, whose architectures are different in only the decoder portion of the network as shown in Figure \ref{fig:Decoder }. 

FCN-32s: Decoder with just one upsampling step of stride 32 for the final layer to recover the predictions for every pixel in the original image. 

FCN-16s: Decoder which combines predictions from both the final layer and the pool4 layer, at stride 16, that results in predictions with finer details, while retaining high-level semantic information.

FCN-8s: Decoder which further combines additional predictions from pool3, at stride 8. This provides further precision compared to both FCN-32 and FCN-16 decoder networks.

Memory estimates for the activations and parameters of all layers in encoder and decoder (FCN-8s) are shown in Table \ref{tab:encoder_memory} and Table \ref{tab:decoder_memory} respectively. It can be observed that convolutional layers i.e shallower layers take up a lot of memory for activation, while deeper layers take up a lot of memory for parameters. Also, it is evident that the resources consumed by the decoder network, relative to the encoder network are relatively meager. Thus employing a deeper or complex decoder network might be a good option to improve the accuracy of dense predictions, which was demonstrated in \cite{noh}. These estimates are also used to identify the correct batch size so that the model fits on the computing system during training, which is discussed in the implementation section. 

\section{Implementation}
\subsection{Dataset}
The dataset that is used in this study is the Cityscapes dataset\cite{cordts}. The dataset has 5000 densely annotated images that are split into training, validation, and testing sets as $2975$, $500$, and $1525$ images respectively. The annotations have a total of $35$ classes belonging to $8$ groups as shown in Table \ref{tab:cityscapestable} on pixel-level segmentation and only fine annotated images without any additional training data or augmentations are used in this evaluation. 

\begin{table}[!hbt]
\begin{threeparttable}
\begin{tabular}{|c|c|}
\hline
Group	&  Classes \\
\hline \hline
flat	&	road, sidewalk,\\ & parking$^+$, rail track$^+$ \\
\hline
human	&	person$^*$, rider$^*$ \\
\hline
vehicle	&	car$^*$, truck$^*$, \\& bus$^*$, on rails$^*$,\\& motorcycle$^*$, bicycle$^*$,\\& caravan$^{*+}$, trailer$^{*+}$\\
\hline
construction	&	building, wall,\\& fence, guard rail$^+$, bridge$^+$, tunnel$^+$ \\
\hline
object	&	pole, pole group$^+$,\\& traffic sign, traffic light\\
\hline
nature	& vegetation terrain\\
\hline
sky	& sky \\
\hline
void	& ground$^+$, dynamic$^+$, static$^+$ \\
\hline        
\end{tabular}
\begin{tablenotes}\footnotesize
\item[*] Will be labeled as group if the boundary between such instances cannot be clearly seen.
\item[+] This label is not included in any evaluation and treated as void.
\end{tablenotes}
\end{threeparttable}
\caption{Cityscapes Dataset: Groups and Classes}
\label{tab:cityscapestable}
\end{table}
The images in the original dataset have a resolution of 2048 x 1024. Using full size images is beyond the compute capability of the system that was used as demonstrated by the memory requirements in Table \ref{tab:encoder_memory}. However, the original images are scaled down to 256 x 512 for training/validation/testing and scaled back up using \textit{INTER NEAREST} interpolation in OpenCV\cite{opencv}. Code from \cite{cordts} was used to pre-process the data to use all labeled classes and generate ground truth images for updated labels as well as validation and calculating the metrics. 



    


\subsection{Hardware and Software}
\begin{table*}[!hbt]
\begin{center}
\begin{tabular}{|c|c|c|c|c|c|}
\hline
Device & Processor & RAM & NVIDIA GPU & VRAM & Compute Capability \\ \hline \hline
Desktop & i7-3770K 8 cores & 32 GB & 2 x GTX1080s & 16 GB & 6.1 \\ \hline
Laptop & i7-7700HQ 4 cores & 32 GB & GTX 1050 & 4 GB & 6.1 \\ \hline
Jetson TX1 & Quad ARM A57 4 cores & 4 GB shared & \begin{tabular}[c]{@{}c@{}}NVIDIA Maxwell\\  256 CUDA cores\end{tabular} & 4 GB  & 5.3 \\ \hline
\end{tabular}
\caption{Hardware Setup}
\label{hwsetup}
\end{center}
\end{table*}


The hardware setup across the devices is as described in Table. \ref{hwsetup}. The desktop and laptop setups have x86\_64 Intel architectures while the Jetson TX1 has an ARM64 processor. The software setup across all devices is attempted to be kept constant to achieve an accurate comparison in their inference time and power parameters. All devices run an Ubuntu 16.04 Linux distro with the exception that the OS for the Jetson TX1 is optimized and packaged within the Jetpack 3.2 package as L4T from NVIDIA. Tensorflow 1.8.0 is installed on all devices, available as prepackaged wheel packages for x86\_64 architectures and had to be compiled from source to run on the ARM64 processor on the Jetson TX1. CUDA 9.0 libraries and cuDNN are installed across all devices. Desktop and laptop builds have complete installations of TensorRT 4.0.0.3, while Jetpack 3.2 currently supports TensorRT 3.0 RC without Python API support. All other dependencies are met using either prepackaged installers or compiled from source for the ARM architecture which proved to be a cumbersome task on the weaker ARM processor.



\subsection{Training}

\subsubsection{Learning Rate}
A parameter search with different learning rates is performed for about 10 epochs on the FCN-8s network and determined that an initial learning rate of 0.0001 and reducing it to 0.00001, after about 25 epoch, works best. No such study was performed for FCN-16s and FCN-32s networks, however, the learning rate for FCN-16s and FCN-32s are increased by a factor of 10 and 100 respectively, which seemed to work well. 

\subsubsection{Batch Size}
Based on the memory calculation shown in Table \ref{tab:encoder_memory} and Table \ref{tab:decoder_memory}, The memory consumption for each image is estimated to be about $170$ MB $*$ $2$ (forward and backward pass) + $135$ MB $*$ $4$ (to account for Adam optimization) = $~1$GB. Since the system used for training has a memory of $16$ GB, a batch size of $10$ is used.

\subsubsection{Cross Entropy Loss (CE Loss)}
Cross-entropy loss is given by the following equation \cite{cl}.
\begin{eqnarray}
loss(f) = −\frac{1}{p}\sum_{i=1}^p \log f_i(y_i^*)
\end{eqnarray}
where $p$ the number of pixels in the image or batch considered, $y_i^* \epsilon \mathit{C}$ the ground truth class of pixel $i$, $f_i(y_i^*)$ the network probability estimate of the ground truth probability of pixel $i$, and $f$ a vector of all network outputs $f_i(c)$, which is obtained by mapping the unnormalized scores $F_i(c)$ of the network through a softmax function.
\begin{eqnarray}
f_i(c) = \frac{e^{F_i(c)}} {\sum_{c^\prime \epsilon \mathit{C}} e^{F_i(c^\prime)}} \quad \forall i \epsilon [1, p], \quad \forall c \epsilon \mathit{C}. 
\end{eqnarray}

The three FCN (FCN-8s, FCN-16s, and FCN-32s) networks are implemented using TensorFlow. The encoder network is initialized with pre-trained VGG16 weights provided by Udacity and trained end-to-end including the encoder for about $50$ epochs each using Adam optimization with minimizing cross-entropy loss as the goal. 

\section{Experiments}

\subsection{Metric}
As the performance measure, the commonly used intersection over union metric will be used, which is evaluated for individual classes
and categories. It is the standard Jaccard Index,
commonly known as the PASCAL VOC intersection-over-union
metric $IoU =
\frac{TP}{TP+FP+FN}$
\cite{pascalVOC}, where $TP$, $FP$, and $FN$
are the numbers of true positive, false positive, and false
negative pixels, respectively, determined over the whole test
set. 



\begin{table}[]
\centering
\begin{tabular}{l|c|c|}
\cline{2-3}
                            & CE Loss                  & mean IOU                   \\ \cline{2-3} 
\multicolumn{1}{c|}{FCN8s}  &\includegraphics[width=2.5cm, height=1.56cm]{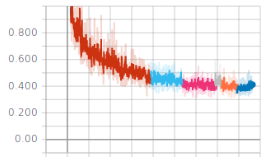}                       &        \includegraphics[width=2.5cm, height=1.56cm]{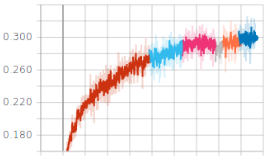}               \\ \cline{2-3} 
\multicolumn{1}{c|}{FCN16s} & \includegraphics[width=2.5cm, height=1.56cm]{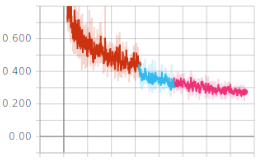}                      &             \includegraphics[width=2.5cm, height=1.56cm]{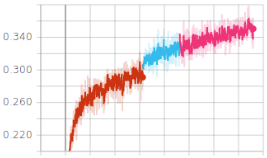}          \\ \cline{2-3} 
FCN32s                      & \multicolumn{1}{l|}{\includegraphics[width=2.5cm, height=1.56cm]{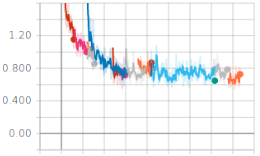}} & \multicolumn{1}{l|}{\includegraphics[width=2.5cm, height=1.56cm]{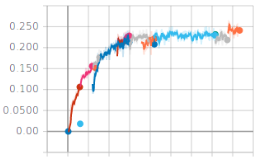}} \\ \cline{2-3} 
\end{tabular}
\caption{Training Characteristics}
\label{trainchar}
\end{table}

\subsection{Segmentation Performance}
The segmentation performance is evaluated on the validation set using the official Cityscapes evaluation script. A per-class mean IOU of $0.421$, $0.461$ and $0.241$ and per-category mean IOU of $0.696$, $0.709$ and $0.533$ for FCN-8s, FCN-16s and FCN-32s respectively. Detailed per class classification results are presented in Table \ref{classIOU} and per category classification results in Table \ref{categoryIOU}. 

The qualitative results of the network are presented in Table \ref{segmentationmaps}. Also, it can be noticed that the network tends to fail in labeling too large or small objects, due to its fixed-size receptive field \cite{noh}. This trend can also be seen in the poor per-class IOU performance on the pole, traffic light or motorcycle classes, for example.

The learning curves for all the networks as well the mean IOU for training set as a function of training epochs is shown in Table \ref{trainchar}. Due to the time constraints of the project, the training had to be stopped after about $50$ epochs each for all the networks, before the networks have fully attained its capacity. This explains slightly better performance for FCN-16s compared to FCN-8s and the lack of finer object structures in images for FCN-32s network relative to the other two networks. For example, in Image4 in Table \ref{segmentationmaps}, traffic signs and traffic lights are completely missing in FCN-32s.  Also, comparing this data with the benchmark data for Cityscapes test set shown in Table \ref{benchmarks} for two networks, which are designed with the goal of reducing inference time, the segmentation results obtained in this study are about $15-20$ \% lower, which can also be understood, given the network training was stopped prematurely.

\begin{table}[h]
\centering
\begin{tabular}{|c|c|c|c|}
\hline
classes  &  FCN-8s  & FCN-16s &    FCN-32s  \\ \hline\hline
road      &  0.94 & 0.942   &  0.918\\ \hline 
sidewalk    & 0.635  &0.657   & 0.505 \\  \hline 
building     & 0.809  &0.811  & 0.726 \\ \hline
wall          & 0.238  &0.209  & 0.0 \\  \hline  
fence         &  0.195 &0.201  &  0.0 \\ \hline 
pole         &  0.171 &0.213   &  0.0\\ \hline
traffic light & 0.111  &0.159  &  0.0\\ \hline
traffic sign  & 0.295  &0.356   & 0.0\\ \hline 
vegetation    & 0.841  &0.836  &  0.738\\  \hline
terrain       & 0.453  &0.447 &   0.233\\  \hline
sky           &  0.870 &0.865  &  0.770\\ \hline
person        & 0.464  &0.486  &  0.016\\ \hline
rider         &  0.035 &0.166  &  0.0\\\hline
car           &  0.832 &0.837 &   0.683\\\hline
truck         & 0.203  &0.330  &  0.0\\\hline
bus           &  0.360 &0.453  &  0.0\\\hline
train         &  0.177 &0.221  &  0.0\\\hline
motorcycle    & 0.061  &0.120  &  0.0\\\hline
bicycle       &  0.452 &0.443  &  0.0\\\hline\hline
Score Average &	0.428&0.461 & 0.241\\\hline
\end{tabular}
\caption{Class IOU}
\label{classIOU}
\end{table}

\begin{table}[h]
\centering
\begin{tabular}{|c|c|c|c|}
\hline
Category  &   FCN-8s & FCN-16s &    FCN-32s  \\ \hline\hline
construction     & 0.792  & 0.797    & 0.702 \\ \hline 
flat    &  0.935 &0.936   &  0.908 \\  \hline 
human     &  0.457 &0.484  &  0.014\\ \hline
nature          & 0.835  &0.833  & 0.726\\  \hline  
object        &  0.206 &0.258  &  0.0\\ \hline 
vehicle        & 0.779  &0.790   &  0.610\\ \hline
sky & 0.870  &0.865  &  0.770 \\ \hline \hline
Score Average &0.696	&0.709 & 0.533 \\\hline
\end{tabular}
\caption{Category IOU}
\label{categoryIOU}

\end{table}

\begin{table}[h]
\centering
\begin{tabular}{|c|c|c|}
\hline
 &   class IOU & Category IOU  \\ \hline\hline
SqueezeNet&& \\based network\cite{speeding}    & 0.598  & 0.843  \\ \hline 
ENet \cite{enet}   &  0.583 &0.804  \\  \hline 
\end{tabular}
\caption{Benchmarks - mean IOU for Cityscapes test set}
\label{benchmarks}

\end{table}

\begin{table*}
\centering
\begin{threeparttable}
\begin{tabular}{|c|c|c|c|c|}
\hline
 & Image1 & Image2 & Image3 & Image4 \\ \hline
Original & \includegraphics[width=3.3cm, height=2cm]{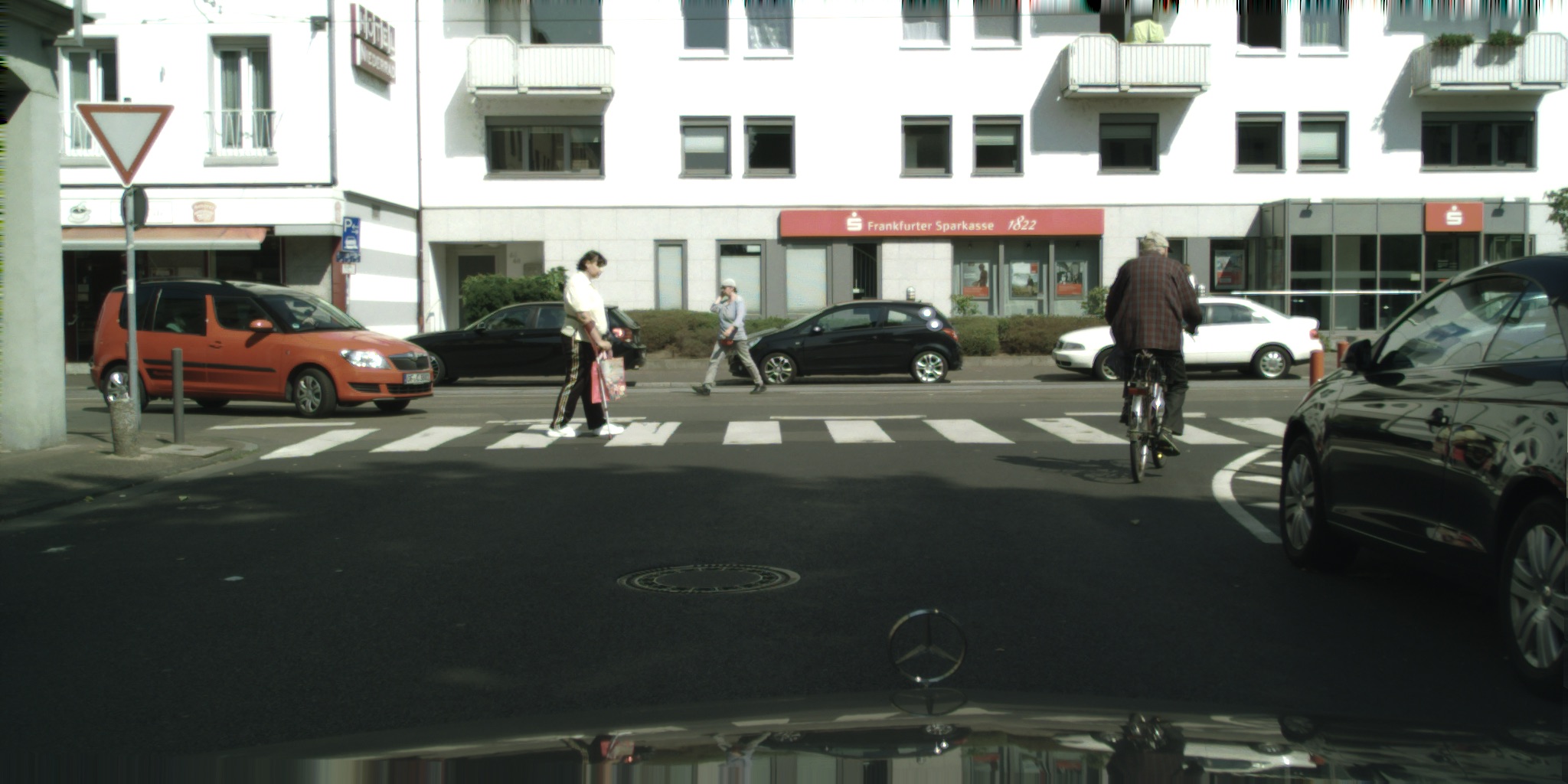} & \includegraphics[width=3.3cm, height=2cm]{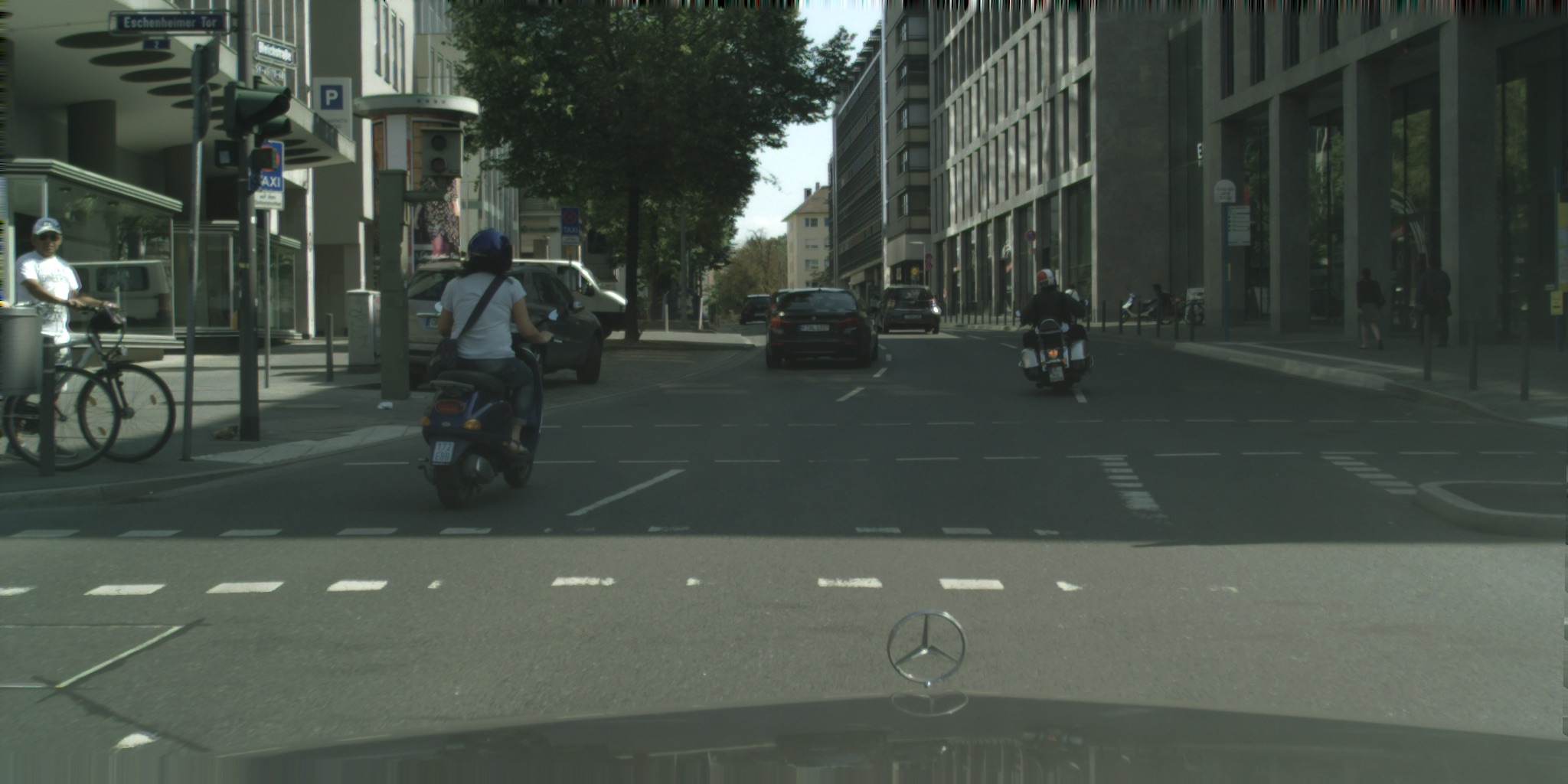}  & \includegraphics[width=3.3cm, height=2cm]{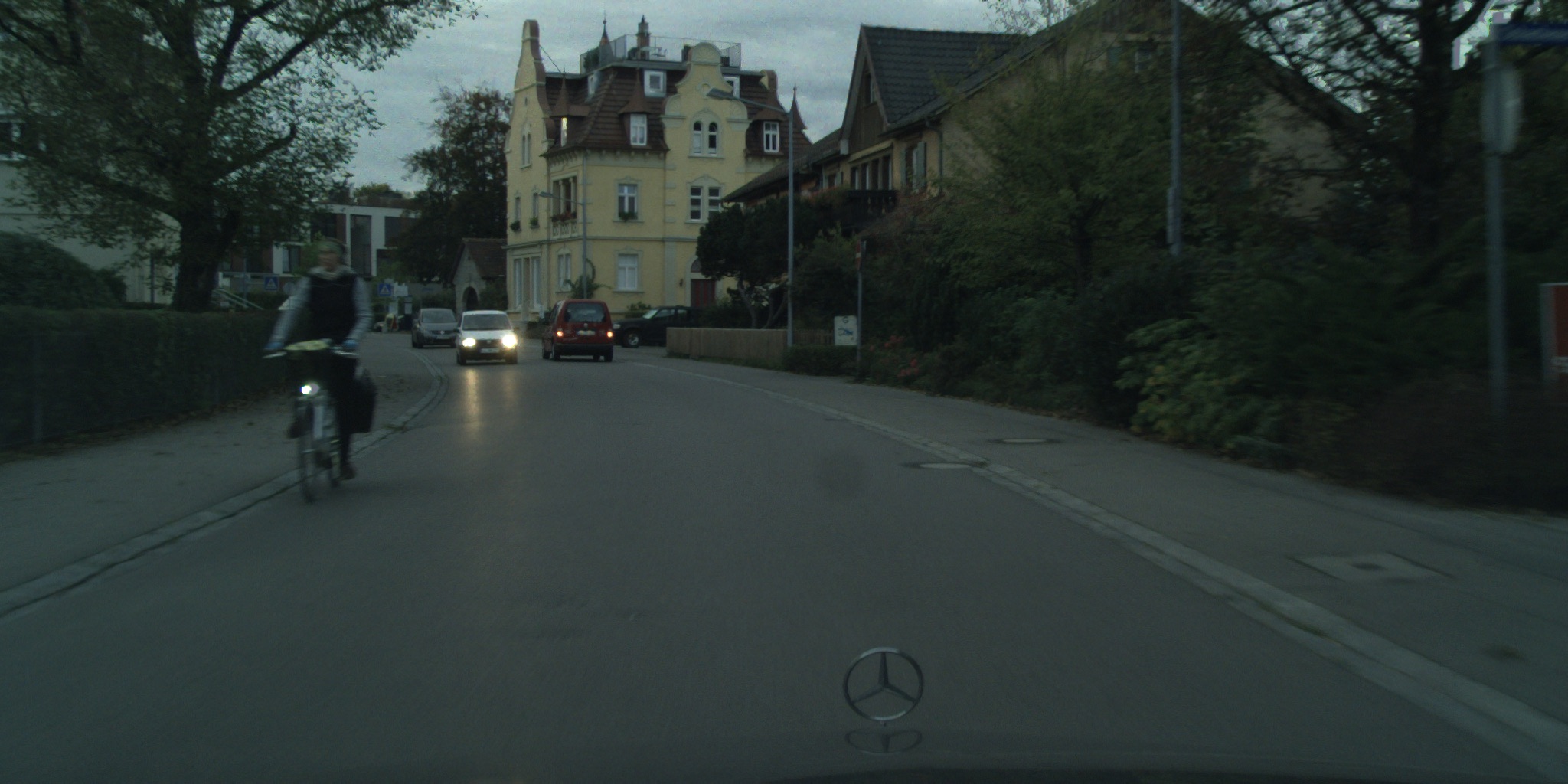}  &  \includegraphics[width=3.3cm, height=2cm]{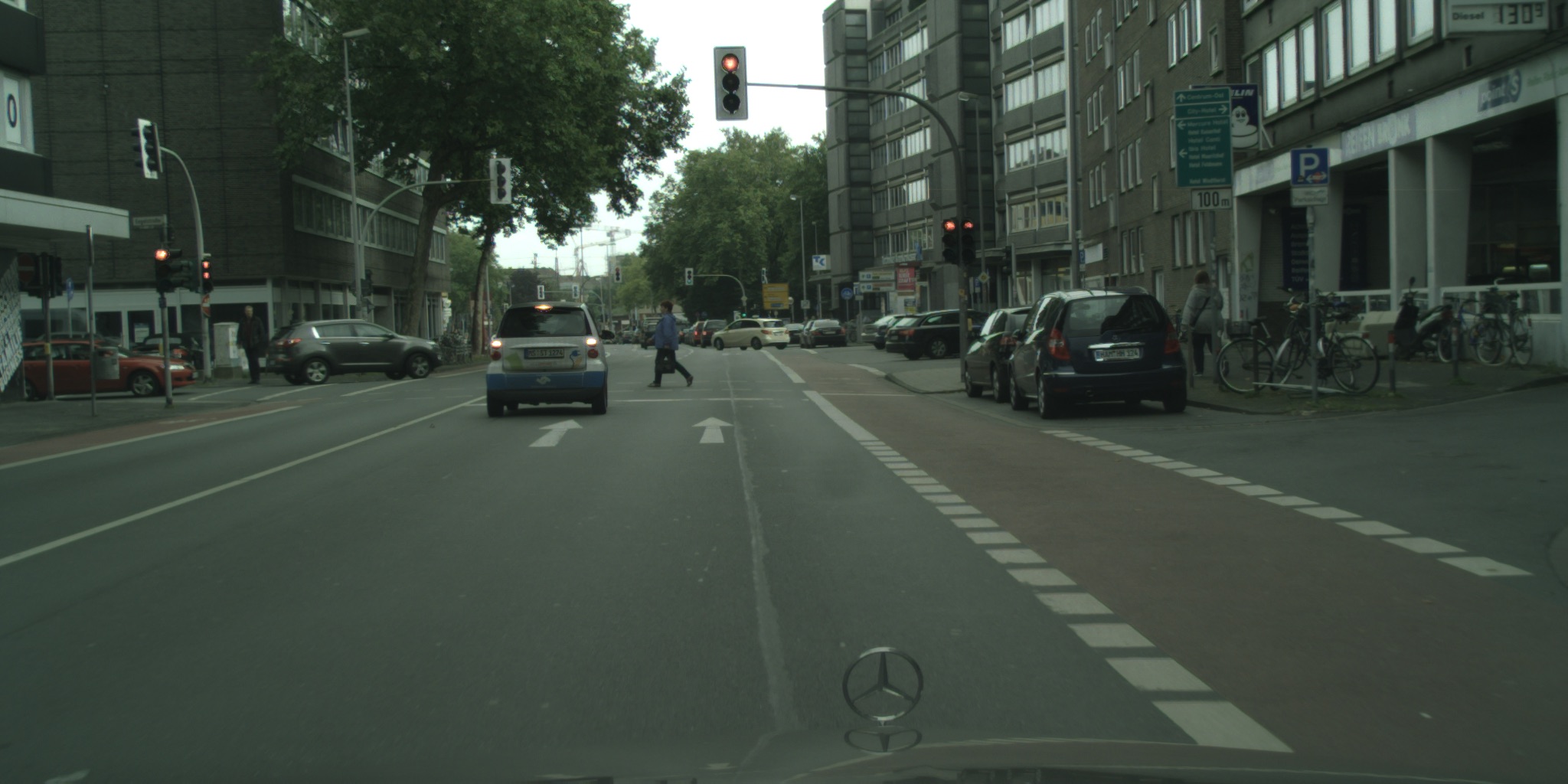}\\ \hline
GT\tnote{+} & \includegraphics[width=3.3cm, height=2cm]{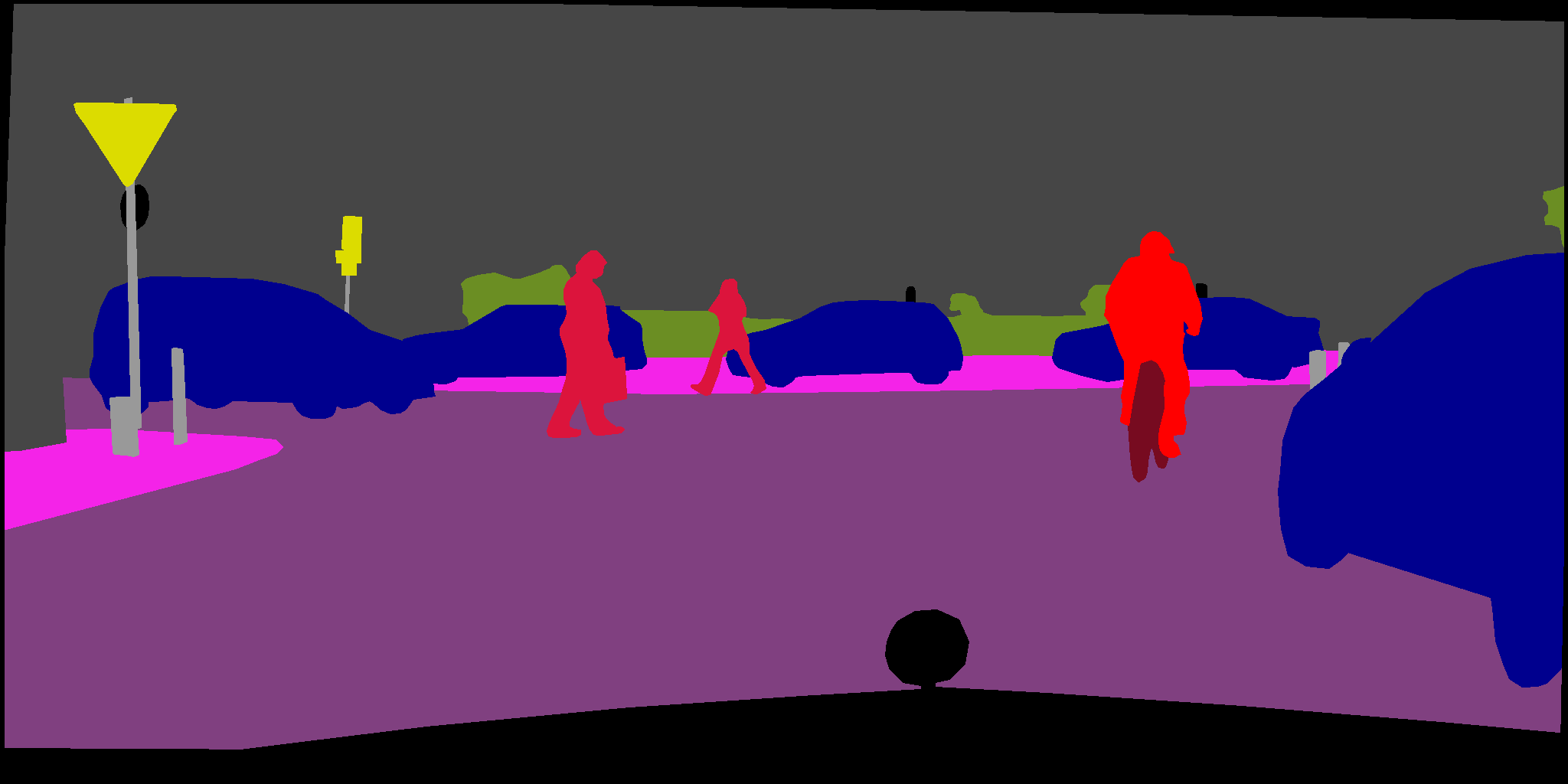} & \includegraphics[width=3.3cm, height=2cm]{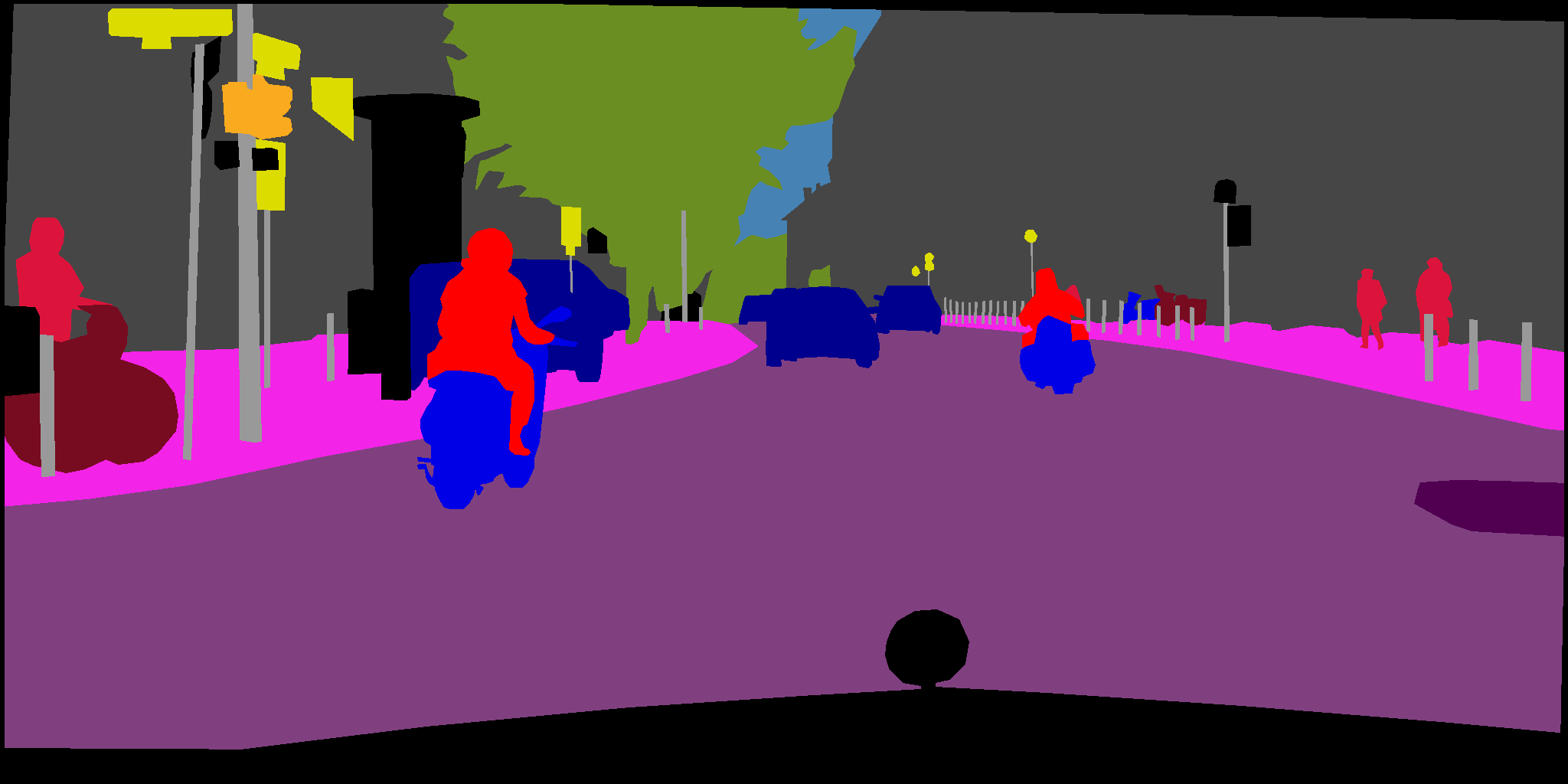}  & \includegraphics[width=3.3cm, height=2cm]{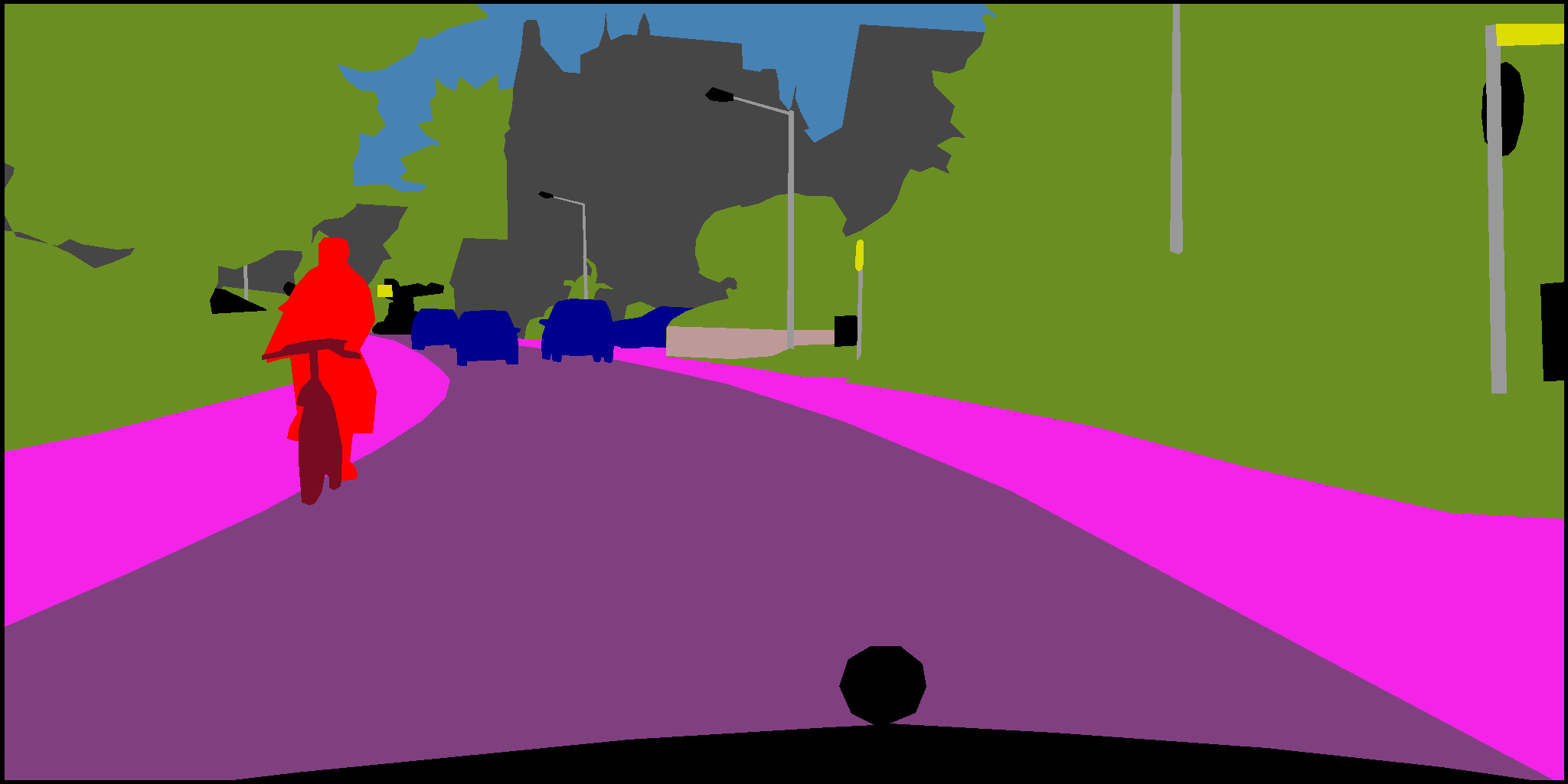}  &  \includegraphics[width=3.3cm, height=2cm]{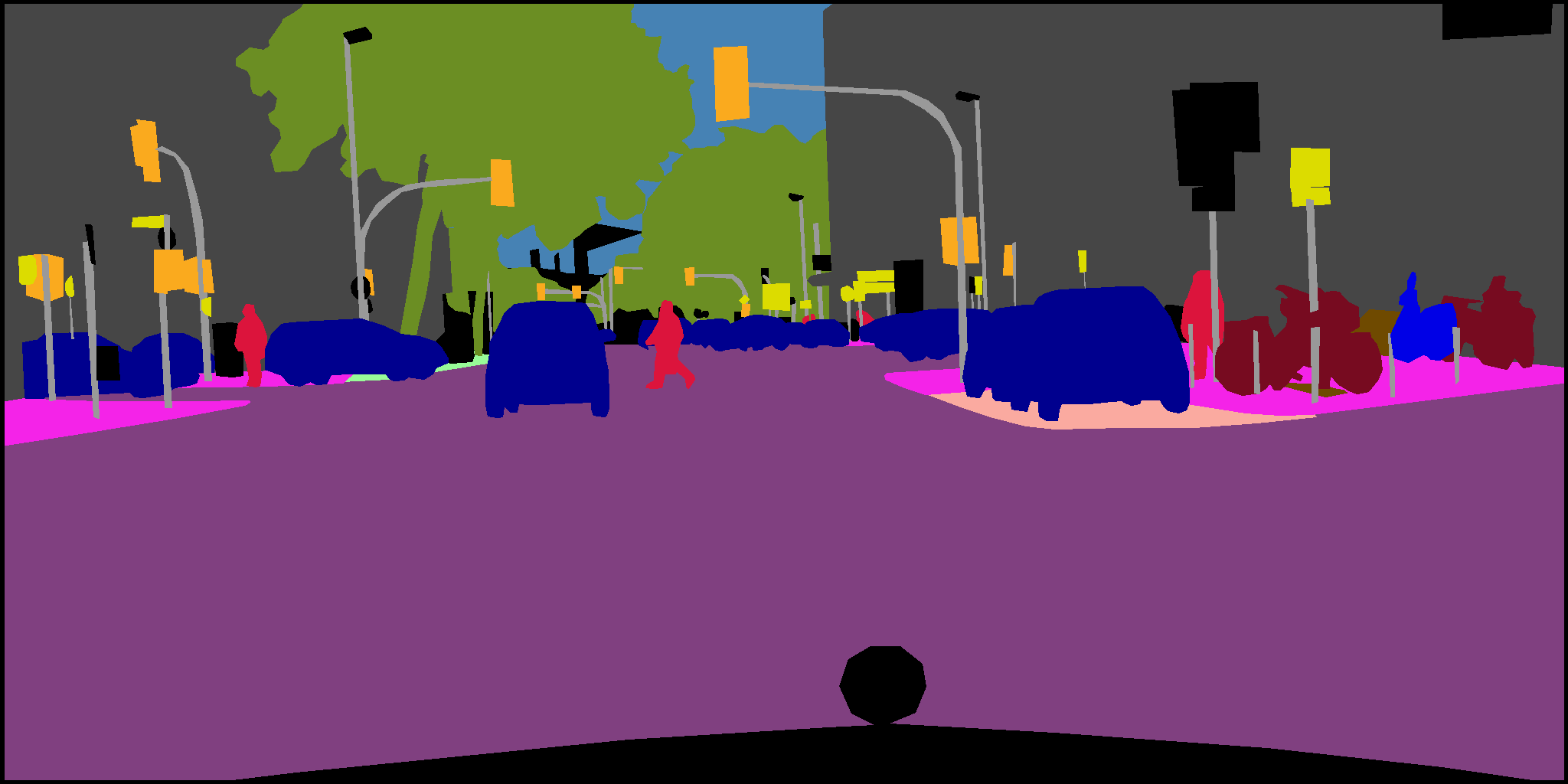}\\ \hline
FCN8s & \includegraphics[width=3.3cm, height=2cm]{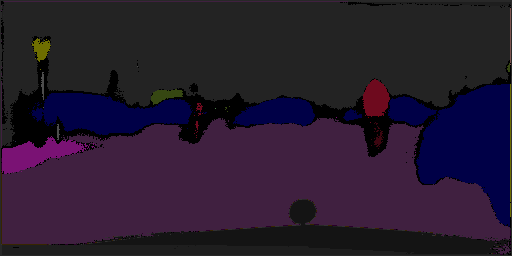} & \includegraphics[width=3.3cm, height=2cm]{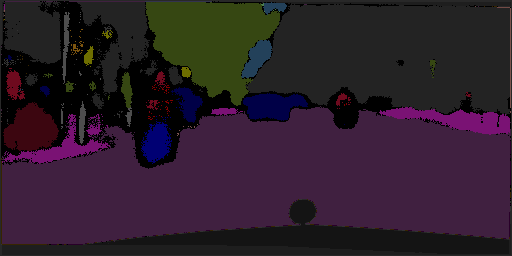} & \includegraphics[width=3.3cm, height=2cm]{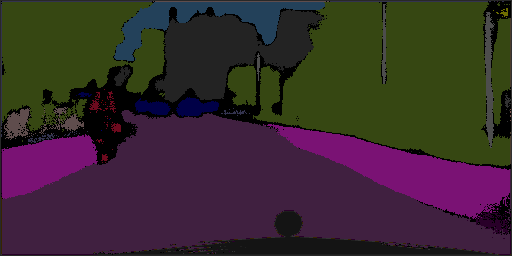} & \includegraphics[width=3.3cm, height=2cm]{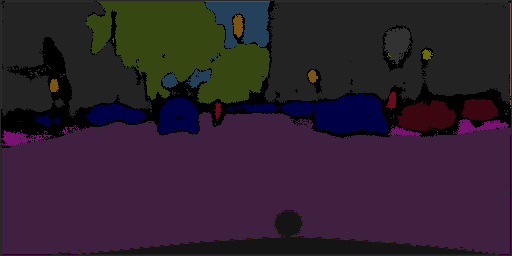}\\ \hline
FCN16s & \includegraphics[width=3.3cm, height=2cm]{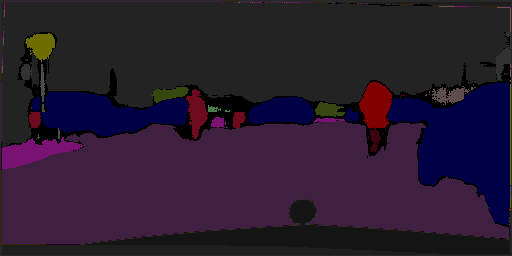} & \includegraphics[width=3.3cm, height=2cm]{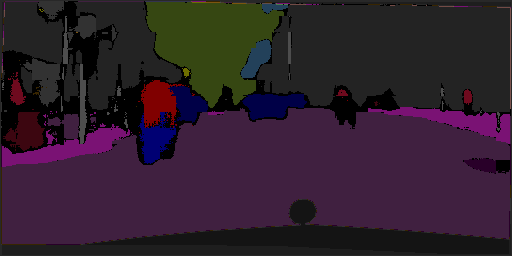} & \includegraphics[width=3.3cm, height=2cm]{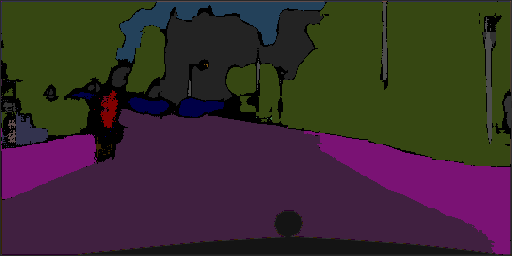} & \includegraphics[width=3.3cm, height=2cm]{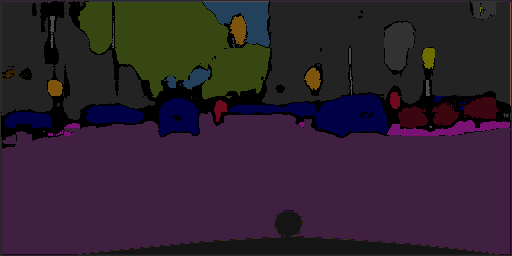}\\ \hline
FCN32s  & \includegraphics[width=3.3cm, height=2cm]{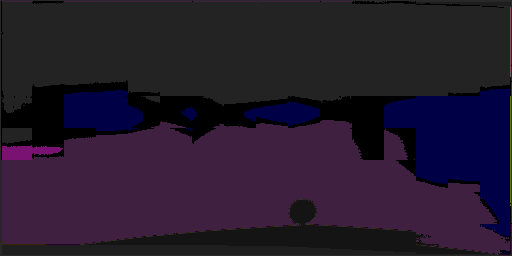} & \includegraphics[width=3.3cm, height=2cm]{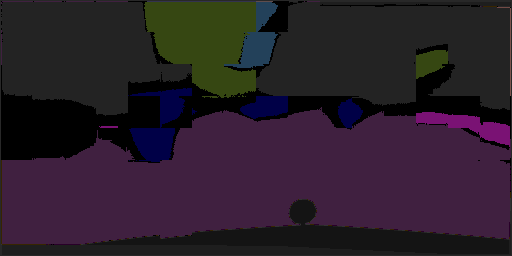} & \includegraphics[width=3.3cm, height=2cm]{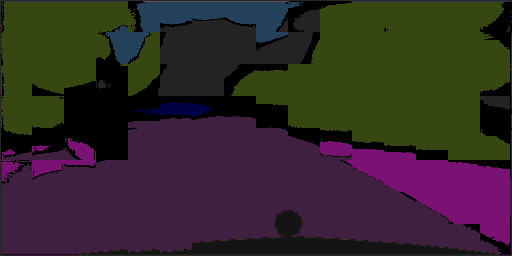} & \includegraphics[width=3.3cm, height=2cm]{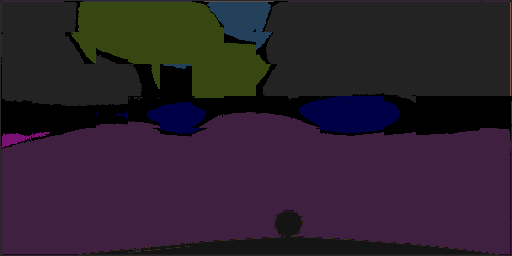}\\ \hline
\end{tabular}
\begin{tablenotes}\footnotesize
\item[+] Ground Truth
\end{tablenotes}
\end{threeparttable}
\caption{Semantic Segmentation Inference Maps }
\label{segmentationmaps}
\end{table*}

\subsection{Effect of built-in TensorFlow optimizations on model size and inference time}
\begin{table*}[]
\centering
\begin{tabular}{|c|c|c|c|}
\hline
\multirow{2}{*}{\textbf{Optimization Parameter}} & \multicolumn{3}{c|}{\textbf{Model Size in MB}} \\ \cline{2-4} 
 & \textbf{FCN8s} & \textbf{FCN16s} & \textbf{FCN32s} \\ \hline
frozen model no optimizations & 153.715 & 154.497  & 140.013 \\ \hline
add\_default\_attributes & 153.715 & 154.497 & 140.013 \\ \hline
fold\_constants(ignore\_errors=true) & 153.715 & 154.497 & 140.013 \\ \hline
fold\_batch\_norms & 153.715 & 154.497 &  140.013 \\ \hline
fold\_old\_batch\_norms & 153.715 & 154.497  & 140.013  \\ \hline
fuse\_resize\_and\_conv & 153.715 & 154.497 &  140.013 \\ \hline
quantize\_weights & 38.481 & 38.673 & 35.048 \\ \hline
strip\_unused\_nodes & 153.715 & 154.497 & 140.013  \\ \hline
sort\_by\_execution\_order & 153.715 & 154.497 &  140.013 \\ \hline
remove\_nodes(op=Identity, op=CheckNumerics) & 153.709 & 154.493 &  140.008 \\ \hline
merge\_duplicate\_nodes & 153.710 & 154.494  &  140.010\\ \hline
All Optimizations & 38.467 & 38.670 & 35.039 \\ \hline
\end{tabular}
\caption{Optimization Parameters and Network Size}
\label{opnns}
\end{table*}


\begin{table*}[]
\centering
\begin{tabular}{|c|c|c|c|c|c|c|c|c|c|}
\hline
\multirow{2}{*}{\begin{tabular}[c]{@{}c@{}}Inference \\ Time (ms) =\textgreater{} \end{tabular}} & \multicolumn{3}{c|}{Desktop} & \multicolumn{3}{c|}{Laptop} & \multicolumn{3}{c|}{Jetson TX1} \\ \cline{2-10} 
                                                       & FCN8s  & FCN16s  & FCN32s & FCN8s  & FCN16s & FCN32s & FCN8s   & FCN16s   & FCN32s  \\ \hline \hline
Baseline                                               & 26      & 24       & 23      & 75      & 69      & 64      & -        & -         & -        \\ \hline
Weight Quantized                                       & 60      & 60       & 56      & 95      & 90      & 82      & 760      & 772       & 714      \\ \hline
\end{tabular}
\caption{Effect of TensorFlow optimizations on inference time (ms)}
\label{inftime}
\end{table*}





A promising approach to reducing inference time and DRAM footprint (power consumption) is model compression. A compressed model that can easily fit into on-chip SRAM cache rather than off-chip DRAM memory will facilitate the deployment of deep networks in self-driving cars where memory size, inference speed, and network bandwidth are all strictly constrained. These will enable a fully trained network to be loaded into SRAM of an embedded processor inside a driverless car, thus providing on-chip in-memory inference at low power \cite{han}. Therefore, the effect of the optimization techniques in the Graph transform tools in TensorFlow \cite{tensorflow2015-whitepaper} on the model size as well as the inference time is quantified in this study.

The first step in deploying a trained network is to freeze the network i.e fuse the information stored in graph definition and checkpoint files by fixing the weights of the network and removing irrelevant training information such as the optimizer options, gradients, etc. During training, weights are not stored in graph definitions as they are constantly tuned and are hence stored in separate checkpoint files, freezing removes the overhead incurred in fetching the latest variable values from separate files.


Once the network is frozen, the TensorFlow provided Graph transform tool can be used to perform optimizations on the saved GraphDef files(.pb). The transform graph tool supports a variety of transforms that can be applied to a network to optimize its size. The optimizations suggested in TernsorFlow's documentation for deployment include stripping unnecessary and unused nodes, folding constants and batch norms and quantizing weights. Table ~\ref{opnns} shows the various optimizations performed on the graph model and their effect on the model size. It is evident that the model size essentially remains the same for the majority of optimizations within an order of a few bytes, except for weight quantization, that converts large floating constant ops into 8-bit equivalents, where the model size reduced to ~$1/4$th of the original size. 



Table ~\ref{inftime} gives the impact of the optimizations on the inference time of the graph. As expected, the optimizations which did not impact the model size also did not cause any significant changes in the inference times, and hence are grouped together with baseline inference time. Weight quantization, on the other hand, has a drastic effect on the inference time. Interestingly, this did not result in the reduction of inference times but rather increased by a factor of $2$. This might be either due to the additional operations that are needed to work with quantized weights or due to the lack of system level drivers that leverage the memory optimizations mentioned in \cite{han}. 


The effects of the underlying hardware platforms on the inference times are also quantified in Table ~\ref{inftime}. As expected, the inference time varies inversely proportional to the compute capability of the hardware, which is estimated to be of the order of $20:3:1$ for desktop, laptop, and Jetson TX1 respectively based on pure compute capability. (The NVIDIA Jetson TX1 has 256 CUDA cores and shared RAM, while the laptop has a GTX1050M GPU with 4GB DDR5 VRAM with 768 CUDA cores, which is three times more than the TX1 and the desktop setup has two NVIDIA GTX1080s with a total of 16GB RAM and 5120 CUDA cores (2560 x 2)  which are about ~20 times more than TX1 and ~7 times more than the laptop.) In reality, this came to around $760:95:60$ $=$ $13:1.5:1$, which is reasonably close, at least in order, to the theoretical estimate. When the baseline model is run on the TX1, an OOM(out of memory) error occurs and the process is killed, hence baseline results are not available. Weight quantized model, however, runs on the TX1 with no problems, which accounts for a need of such optimizations for embedded platforms.

\subsection{Effect of built-in TensorRT optimizations on inference time}

employing optimizations and calibrations to neural networks to obtain optimal performance in GPUs designed by NVIDIA. The effect of TensorRT optimizations on the inference times of the network is quantified in this study as the testing platforms are based on NVIDIA GPUs. 

TensorFlow graphs are exported for use with other backends using the Universal Framework Format(uff), when a uff graph is parsed into a TensorRT engine the following four automatic optimizations are performed. Layer and Tensor fusion reduce the number of layers by recognizing and fusing layers that have the same input data and filter size, and CUDA kernels are fused together to perform sequential operations to overcome latency introduced due to multiple kernel launches. Precision calibration allows choosing the inference precision between FP32, FP16 or INT8 without a need for retraining the network. Kernel auto-tuning chooses an optimized kernel from a wide range of options to best suit the target GPU, input data, batch size, tensor layout, and many other such parameters. Dynamic Tensor Memory ensures that memory is reused by designating memory for a tensor only while it is being used, which prevents memory allocation overhead. These optimizations should bring a significant reduction in the inference time of the networks.

The bonnet framework \cite{bonnet} provides a C++ API for TensorRT, which is used to run the TensorRT optimizations on the three platforms and their effect on the inference time is documented in Table ~\ref{trtop}. As expected TensorRT optimizations showed a significant reduction in inference times with a reduction of ~$33$\% on the desktop and ~50\% on the laptop. 



\begin{table}[]
\centering
\begin{tabular}{|c|c|c|}
\hline
Device     & Baseline (ms) & TensorRT optimized (ms)\\ \hline
Desktop    & 26          & 9        \\ \hline
Laptop     & 75         & 34         \\ \hline
Jetson TX1 & -             & 460         \\ \hline
\end{tabular}
\caption{Effect of TensorRT optimizations on inference time (ms)}
\label{trtop}
\end{table}


\subsection{Comparison of performance and power metrics across hardware platforms}

Given that three hardware platforms have different compute capability as well as power consumption, to quantify and compare the inference times across platforms, the inference times should be normalized by the power consumed.

The power consumption data for the desktop and laptop devices are measured using the NVIDIA System Management Interface available as a command line utility with relevant parameters as shown below. \\
\footnotesize
\verb`nvidia-smi daemon -i 0 -s p -d 5 -p /data/logs`\\
\normalsize
The power reading provided by the tool is measured in Watts(W) for each GPU and is accurate to +/- 5 watts.

Measuring power on the Jetson TX1 is not as straightforward as the custom graphic driver is not bundled with SMI. The TX1 has INA monitors to measure current and voltage being drawn and are available to the processor through an i2c interface. The TX1 has a three channel monitor which provides the input current(mA), voltage(mV) and  power(mW) at i2c address 0x40. The commands required to obtain these readings are:

\noindent \footnotesize \verb`cd /sys/devices/platform/7000c400.i2c/i2c-1/`
\verb`cat 1-0040/iio_device/in_current0_input`
\verb`cat 1-0040/iio_device/in_voltage0_input`
\verb`cat 1-0040/iio_device/in_power0_input`

\normalsize



The outputs of the \verb`cat` command can be redirected to a file for processing. Table \ref{pow} shows the average power consumed by each of the hardware platforms. Energy consumption, E(W-Hr) is the energy consumed by the platform to run $1525$ test images, which is given in terms of inference time (ms) and average power consumed (P).
\begin{equation}
E = \frac{1525 * I * P}{60 * 60 * 1000} \nonumber
\end{equation}
The relative performance of the network on the three platforms when measured purely in regards to inference times, i.e how many images can be inferred per second is $51:4:1$ (inversely proportional to inference times). However, when the performance is compared in regards to energy consumed i.e how many images can be inferred per W-hr, it changes to 
$6:2.5:1$, outlining the power efficiency of the embedded system.


\begin{table}[]
\centering
\begin{tabular}{|c|c|c|}
\hline
Device     & Power & Energy\\
&Consumption(W) & Consumption(W-hr) \\ \hline \hline
Desktop    & 35.27  &  0.134              \\ \hline
Laptop     & 22.65   & 0.326           \\ \hline
Jetson TX1 & 4.16     & 0.810               \\ \hline
\end{tabular}
\caption{Average Power consumption for inference}
\label{pow}
\end{table}

\section{Conclusion and Future Work}
In this project, the task of pixel-wise semantic segmentation in the context of self-driving with a goal to reduce inference time is explored. FCN based networks with a VGG16 encoder architecture and skip connections are trained on the Cityscapes dataset. On the validation set, the trained networks scored a per-class mean IOU class $0.421$ $0.461$ $0.241$ and per-category mean IOU of $0.696$, $0.709$, $0.533$ for FCN-8s, FCN-16s and FCN-32s networks respectively. Several network optimizations built into TensorFlow and TensorRT and their impact on inference times as well as model size are quantified. Finally, the trained network is ported onto Jetson TX1 and inference times across the hardware platforms are compared and presented. 


This work could be further extended in several ways. In this study, though the inference times across hardware platforms are compared, corresponding IOU scores for the validation/test sets are not obtained, which is needed to fully understand the accuracy and inference tradeoff. Networks based on more efficient architectures such as SqueezeNet \cite{squeezenet} coupled with optimizations could also be looked into to quantify their performance metrics on embedded platforms. Also, optimization techniques that need retraining such as pruning are not considered in this experiment, which could be explored as well.


{\small
\bibliographystyle{ieee}
\bibliography{egbib}
}
\end{document}